\documentclass{article}
\usepackage{spconf,amsmath,graphicx}
\usepackage{spconf,amsmath,epsfig}
\usepackage{algorithm}
\usepackage{algorithmic}
\usepackage{hhline}

\usepackage{algorithm}
\usepackage{algorithmic}
\usepackage{graphicx}
\usepackage{amssymb}
\usepackage{booktabs}
\usepackage{amsfonts}

\usepackage{multirow}
\usepackage{array}
\usepackage{color}
\usepackage{subcaption}
\usepackage{colortbl}
\usepackage{arydshln}
\graphicspath{{images/}}
\usepackage{bbding}
\usepackage{makecell}
\usepackage{hhline}
\usepackage{bm}

\title{Improving Image Captioning with Control Signal of Sentence Quality}
\name{Zhangzi Zhu, Shuai Wang, Hong Qu*\thanks{This work was supported the National Science Foundation of China under Grant 61976043 and 62106038; and in part by the Science and technology support program of Sichuan Province under Grant 2022YFG0313. *Corresponding author: Hong Qu (hongqu@uestc.edu.cn).}}
\address{\normalsize{School of Computer Science and Engineering,
University of Electronic Science and Technology of China} \\
\normalsize{zzcaptain@foxmail.com}}

\begin{document}
%\ninept
%
\maketitle
\begin{abstract}
In the dataset of image captioning, each image is aligned with several descriptions. Despite the fact that the quality of these descriptions varies, existing captioning models treat them equally in the training process. In this paper, we propose a new control signal of sentence quality, which is taken as an additional input to the captioning model. By integrating the control signal information, captioning models are aware of the quality level of the target sentences and handle them differently. Moreover, we propose a novel reinforcement training method specially designed for the control signal of sentence quality: Quality-oriented Self-Annotated Training (Q-SAT). Extensive experiments on MSCOCO dataset show that without extra information from ground truth captions, models controlled by the highest quality level outperform baseline models on accuracy-based evaluation metrics, which validates the effectiveness of our proposed methods.
\end{abstract}
\section{Introduction}
\label{sec:intro}

Image captioning \cite{Vinyals2015Show, xu2015show} aims to describe the visual content of an image with natural language. This task requires to combine knowledge from computer vision to extract visual information and natural language processing to generate fluent sentences.  Applications of image captioning include content-based retrieval and visually impaired people assistance.

\begin{figure}[t]
    \centering
    \begin{minipage}{0.36\linewidth}
        \includegraphics[width=0.95\linewidth]{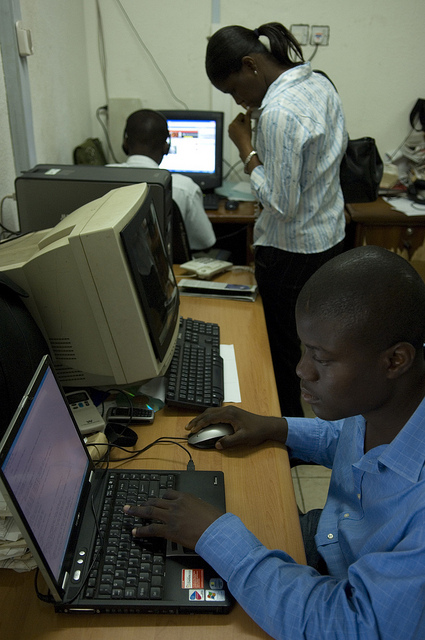}
        \end{minipage}
        \begin{minipage}{0.50\linewidth}
        \footnotesize{
        \textbf{GT1:} Several young students working at a desk with multiple computers.\\
        \textbf{GT2:} A man is working on a laptop next to other computers.\\
        \textbf{GT3:} People in a large room, use multiple computers.\\
        \textbf{GT4:} A young professional is working at his laptop while his coworker is reading material.\\
        \textbf{GT5:} A young man at his workstation examines the monitor of a lap top, with one hand on the keyboard and the other on the mouse.
        }
    \end{minipage}
    \caption{An example of an image with its paired five ground truth captions on MSCOCO dataset. Captions 1-4 are more consistent with mainstream ground truth descriptions while caption 5 has a strong personal style and makes the model convergence difficult.}
    \label{fig1}
\vspace{-.35cm}
\end{figure}

Each image in the dataset of image captioning is usually labeled with several different descriptions. Existing models regard these descriptions as ground truth captions and treat them equally during training. However, they ignore that the quality of such descriptions aligned with the same image is different. Figure \ref{fig1} shows an example from MSCOCO \cite{Lin2014Microsoft} dataset. As we can see, captions 1-4 mainly describe salient objects, their attributes and relationships in the image. They are consistent with mainstream ground truth captions, which are more instructive for the convergence of the captioning model. These captions are classified into `` high quality ''. On the contrary, Descriptions like caption 5 pay more attention to some inconspicuous details in the image. They usually have a strong personal style and make the model convergence difficult, which are classified into `` low quality ''.

Completely discarding such `` low quality '' sentences is a very straightforward method. However, it will greatly hurt the diversity of captioning models. In order to treat descriptions differently according to their quality levels while maintaining the diversity of models, we turn to the field of controllable image captioning. The controllable image captioning (CIC) task aims to generate captions conditioned on designated control signals such as image regions \cite{cornia2019show} or sentence length \cite{deng2020length}. In this paper, we propose a new control signal for CIC tasks: sentence quality. In the training process, captioning models take the sentence quality level as an additional input. By integrating the control signal information into the input, CIC models naturally relate the input quality level with the output sentence, thus learning the meaning of sentence quality. In the evaluation stage, when a specific quality level is determined, the CIC model will review captions of the same level in the training phase and generate sentences of corresponding quality. That is, CIC models are able to generate sentences of different quality levels as needed.

It has been proved in \cite{zhu2021self} that conventional reinforcement training methods are not applicable to CIC tasks. Therefore, we propose a novel reinforcement training method: Quality-oriented Self-Annotated Training (Q-SAT), which is based on Self-Annotated Training (SAT) \cite{zhu2021self} and specially designed for the control signal of sentence quality. Compared to SAT method, we makes some improvements at the beginning to determine the center level of sampled sentences for a given image, which help to reduce the subsequent control signal changes and improve the overall performance of CIC models. Moreover, we design an additional step to retain the optimization of samples resulting in lower rewards than baseline, which is discarded in SAT method since it leads to instability in the training process. Extensive experiments conducted on MSCOCO dataset \cite{Lin2014Microsoft} demonstrate that, without additional information from ground truth captions, models controlled by the highest quality level outperform baseline models on accuracy-based evaluation metrics. The improvement of models trained with cross entropy loss or Q-SAT method validates the effectiveness of our proposed control signal of sentence quality and Q-SAT method.

\section{Method}
\subsection{Embedding Method}
\label{sec3.1}

In this work, we divide captions into several levels according to their quality. Since sentence quality is an abstract concept, we define the quality of sentence $ Y $ as the CIDEr \cite{Vedantam2015CIDEr} score computed with all ground truth captions. The specific settings of control levels related to its CIDEr score during cross entropy training and reinforcement training \cite{rennie2017self} are shown in Table \ref{Table2} respectively. We try to make the number of samples at each level as evenly distributed as possible.

Given an input caption $ Y = \{y_1, y_2, ..., y_T \} $, we first calculate its CIDEr score and assign Y into a specific level $ \beta $ according to Table \ref{Table2}. Then, the quality level embedding for each word is computed by $ e_\beta = W^T\Pi_\beta $, where $ W^T $ is a quality level embedding matrix and $ \Pi_\beta $ is the one-hot representation of $ \beta $. The final representation $ x_i $ of each word $ y_i $ is constructed by adding the quality level embedding $ e_\beta $ with the word embedding $ e_{y_i} $ and, optionally (for Transformer-based \cite{vaswani2017attention} decoder), its positional embedding $ e_{p_i} $:
\begin{equation}\label{equ1}
x_i = e_\beta + e_{y_i} + e_{p_i}.
\end{equation}
In this way, the quality level information is fused into the word embedding, which enables us to control the quality of generated sentences from the exterior. We can easily achieve this goal by replacing the original word embedding with $ x_i $ as the input of the decoder of captioning models.

\begin{table}[htb]
	\begin{center}
		\begin{tabular}{|c|c||c|c|}
		    \hhline{--||--}
		    \multicolumn{2}{|c||}{XE}&\multicolumn{2}{c|}{RL} \\
			\hhline{--||--}
			$ \beta $ & CIDEr score $ x $ & $ \beta $ & CIDEr score $ x $\\
			\hhline{--||--}
			0 & $ x \leq $ 2.3 & 0 & $ x \leq $ 0.7 \\ \hhline{--||--}
            1 & 2.3 $ < x \leq $ 2.5 & 1 & 0.7 $ < x \leq $ 1.3 \\ \hhline{--||--}
            2 & $ x > $ 2.5 & 2 & $ x > $ 1.3 \\	 \hhline{--||--}
		\end{tabular}
	\end{center}
	\caption{Specific settings of control levels of sentence quality during cross entropy training and reinforcement training.}
\label{Table2}\end{table}

\begin{algorithm}[tb]
\caption{Training Procedure of Q-SAT Method}
\label{alg1}
\textbf{Input}: given image $ I $, paired ground truth captions $ G = \{G_1, G_2, \cdots, G_k\} $\\
\textbf{Output}: controllable captioner $ C $
\begin{algorithmic}[1] %[1] enables line numbers
\FOR{epoch in $ [M, N) $}
\STATE Randomly initialize $ k $ control signal levels $ \beta^{init} = \{\beta_1^{init}, \beta_2^{init}, \cdots, \beta_k^{init}\} $.
\STATE Use Monte-Carlo strategy to sample $ k $ captions $ Y^s = \{Y_1^s, Y_2^s, \cdots, Y_k^s\} $ where $ Y_i^s = C(I, \beta_i^{init}) $
\STATE Calculate the average CIDEr score $ S^{avg} $ based on $ Y^s $
\STATE Calculate the average control signal level $ \beta^{avg} $ based on $ S^{avg} $
\STATE Use Monte-Carlo strategy to sample $ k $ captions $ Y^t = \{Y_1^t, Y_2^t, \cdots, Y_k^t\} $ where $ Y_i^t = C(I, \beta^{avg}) $ and get its corresponding output distribution $ p(Y_i^t|I,\beta^{avg}) $.
\FOR{each caption $ Y_i^t $ }
\STATE Calculate the reward $ r(Y_i^t) $ and the control signal level $ \beta_i^s $ based on $ Y_i^t $.
\STATE Change $ \beta_i^s $ to $ \beta_i^{ns} $ as in Eq. (\ref{equ4}).
\STATE Generate output distributions $ p(Y_i^t|I,\beta_i^{ns}) $.
\STATE Change the output distribution $ p(Y_i^t|I,\beta_i^{ns}) $ to $ p'(Y_i^t|I,\beta_i^{ns}) $ as in Eq. (\ref{equ5}).
\ENDFOR
\STATE Optimize $ C $ with Eq. (\ref{equ6}).
\ENDFOR
\end{algorithmic}
\end{algorithm}

\subsection{Quality-oriented Self-Annotated Training}
\label{sec3.3}
Reinforcement training is crucial to CIC tasks since it can solve the problem of exposure bias and inconsistency between the optimizing function and evaluation metrics. However, it has been proved in \cite{zhu2021self} that conventional reinforcement training are not applicable to most CIC tasks. Therefore, we choose SAT method proposed in \cite{zhu2021self} as the baseline method.

In this paper, we propose a novel reinforcement training method designed for quality signal: Quality-oriented Self-Annotated Training (Q-SAT). Algorithm \ref{alg1} summarizes the entire process of our Q-SAT method. Given a controllable captioner $ C $, an image $ I $ and its $ k $ paired ground truth captions $ G = \{G_1, G_2, \cdots, G_k\} $, we first randomly initialize $ k $ control signal levels $ \beta^{init} = \{\beta_1^{init}, \beta_2^{init}, \cdots, \beta_k^{init}\} $. Then, $ k $ sentences $ Y^s = \{Y_1^s, Y_2^s,$ $ \cdots, Y_k^s\} $ are sampled by the Monte-Carlo strategy with the input of $ I $ and $ \beta^{init} $, where $ Y_i^s = C(I, \beta_i^{init}) $. For each sampled sentence $ Y_i^s $ in $ Y^s $, we calculate the CIDEr score between it and $ G $, and finally obtain the average CIDEr score of $ Y^s $, which corresponds to the quality level $ \beta^{avg} $. The above steps are to determine the center level of the CIC model for image $ I $. It means that sampled sentences probably fall into the quality level $ \beta^{avg} $ for image $ I $ , which can reduce the subsequent control signal changes in algorithm step 9. The ablation study in Section \ref{sec4.4} demonstrates the necessity of this improvement.

\begin{table*}[htbp]
	\begin{center}
		\begin{tabular}{c|cccccc|cccccc}
		    \hline
		    {}&\multicolumn{6}{c|}{Cross Entropy Loss}&\multicolumn{6}{c}{CIDEr Score Optimization} \\
			\hline
			Model & B-1 & B-4 & M & R & C & S & B-1 & B-4 & M & R & C & S  \\ \hline
			Transformer$ ^\dag $ & 75.6 & 35.3 & 27.6 & 56.1 & 112.5 & 20.8 & {80.8} & 39.0 & 29.0 & 58.8 & 129.4 & 22.8 \\
			+ quality (0) & 73.9 & 33.2 & 26.9 & 54.8 & 105.9 & 19.8 & 80.6 & 39.0 & {29.2} & 58.9 & 130.9 & {23.0} \\
			+ quality (1) & 75.0 & 34.2 & 27.3 & 55.5 & 109.5 & 20.3 & 80.7 & {39.2} & {29.2} & {59.0} & 131.0 & {23.0} \\ 
			+ quality (2) & {76.4} & {36.0} & {27.9} & {56.6} & {115.1} & {20.8} & {80.8} & {39.5} & {29.2} & {59.0} & {131.4} & {23.0} \\\hline
		\end{tabular}
	\end{center}
	\caption{Performance comparisons between baseline models controlled by different levels of sentence quality, where B-1, B-4, M, R, C and S are short for BLEU1, BLEU4, METEOR, ROUGE-L, CIDEr-D and SPICE scores respectively. All values are reported as percentage ($ \% $). $ ^\dag $ denotes our trained model based on the publicly available source code.}
\label{Table3}\end{table*}

Following the recursive annotation mechanism \cite{zhu2021self}, we take $ \beta^{avg} $ as input to generate $ k $ sentences $ Y^t = \{Y_1^t, Y_2^t, \cdots, $ $Y_k^t\} $ and obtain their corresponding output distributions $ p(Y_i^t|I,\beta^{avg}) $ with the Monte-Carlo strategy for another iteration. For each sentence $ Y_i^t $, we compute the reward $ r(Y_i^t) $ and the corresponding control level $ \beta_i^s $. During SAT \cite{zhu2021self} method, samples resulting in lower rewards than baseline $ b $ are not optimized because they lead to instability in the training process. In our Q-SAT method, we retain the optimization of these sentences. Since the instability mainly results from suppressing the output probability of sentences which are not sampled by the model itself, we change $ \beta_i^s $ to $ \beta_i^{ns} $:
\begin{equation}\label{equ4}
\beta_i^{ns} = \begin{cases}
\beta_i^s, & r(Y_i^t) \geq b \\
\beta^{avg}, & r(Y_i^t) < b  . \\
\end{cases}
\end{equation}
\noindent After that, the model regards $ Y_i^t $ as the target caption and predicts output distributions conditioned on $ I $ and $ \beta_i^{ns} $, i.e. $ p(Y_i^t|I,\beta_i^{ns}) $ as the process of cross entropy training. Then, We update $ p(Y_i^t|I,\beta_i^{ns}) $ to $ p'(Y_i^t|I,\beta_i^{ns}) $ due to the existence of dropout:
\begin{equation}\label{equ5}
p'(Y_i^t|I,\beta_i^{ns}) = \begin{cases}
p(Y_i^t|I,\beta^{avg}), & \beta^{avg} = \beta_i^{ns} \\
p(Y_i^t|I,\beta_i^{ns}), & \beta^{avg} \neq \beta_i^{ns} .\\
\end{cases}
\end{equation}
Finally, the CIDEr reward $ r(Y_i^t) $ and the distribution $ p'(Y_i^t|I, $ $\beta_i^{ns}) $ are sent into Eq. (\ref{equ6}) to optimize the model:
\begin{equation}\label{equ6}
\nabla_\theta L_{RL}(\theta) = -\frac{1}{k}\sum_{i=1}^{k} (r(Y_i^t) - b) \nabla_\theta log(p'_\theta(Y_i^t|I, \beta_i^{ns})).
\end{equation}

\noindent where $ b = (\sum_{i} r(Y_i^t)) / k $ is the baseline, computed as the mean of the rewards obtained by the sampled captions.

\subsection{Inference}
It should be noted that the goal in this work is different from that in \cite{zhu2021self}. For the sentence attribute signal, \cite{zhu2021self} focuses more on improving controllablility when faced with different control signals. In the evaluation phase, the control level is computed from each ground truth caption and then sent into the CIC model to generate the caption of the same attribute. In this case, the generated caption and the ground truth caption are in one-to-one correspondence, so the final result is calculated between them. By contrast, for the sentence quality signal, extra information from the ground truth captions is not needed in the inference stage. We fix the control level of sentence quality to the highest level, which drives the CIC model to generate high-quality sentences. Section \ref{sec4.5} validates they are better than descriptions generated from regular captioning models. The final performance of the model is computed between the generated caption and all ground truth captions paired with the same image. Therefore, the performance of the CIC model controlled by the sentence quality signal is comparable to the existing regular captioning models, since it does not require extra information from the ground truth captions. Except for the highest quality level, other quality levels are used to improve the diversity of the model. For a given image, our model can generate various descriptions.

\section{Experiments}
\subsection{Datasets and Evaluation Metrics}
We evaluate our proposed method on the MSCOCO dataset \cite{Lin2014Microsoft} and follow the setting of Karpathy's split \cite{karpathy2015deep}. To evaluate the quality of generated captions, we calculate evaluation metrics including BLEU \cite{papineni2002bleu}, ROUGE-L \cite{lin2004rouge}, METEOR \cite{denkowski2014meteor}, CIDEr \cite{Vedantam2015CIDEr} and SPICE \cite{anderson2016spice}.

% \noindent\textbf{Implementation Details.} We follow Transformer \cite{2017Attention} to set its hyper-parameters and train the models. Specifically, we use pretrained Faster-RCNN \cite{Ren2015Faster} to obtain a 2048-dimensional feature vector for each region. For Transformer-based models, we set the dimensionality of each layer to 512, the number of heads to 8. We adopt Adam optimizer to minimize the cross-entropy loss for 15 epochs, and then use reinforcement training with a fixed learning rate of $ 5\times10^{-6} $ for another 15 epochs. We train the models with a batch size of 10 and a beam size of 5.

\subsection{Evaluation on Control Signals of Sentence Quality}
\label{sec4.3}
To prove that our control signal is able to determine the sentence quality, we show the performance of baseline models controlled by different quality levels in Table \ref{Table3}. The vanilla Transformer model is trained with conventional RL method \cite{rennie2017self} while models equipped with the control signal of sentence quality are trained with our Q-SAT method.

\noindent \textbf{Controllability.} In terms of cross entropy training, we can observe that the quality of generated sentences varies greatly between different control levels. As the sentence quality level increases, the CIC model generates higher quality captions. For reinforcement training, there is still a gap in accuracy between the highest and the lowest quality level. It can be seen that between adjacent levels, sentences of high quality level surpass those of low quality level in most evaluation metrics. We notice that the gap of metrics between levels in reinforcement training is not as big as that in cross entropy training. It is caused by the fact that the difference between generated captions from trained models (RL training) is smaller than that between ground truth captions (XE training). The experiment proves that models can be externally controllable through the quality signal.

\noindent \textbf{Accuracy.} Compared with the baseline model, the model controlled by the highest level of sentence quality performs much better in accuracy. In terms of cross entropy training, the baseline model equipped with the highest quality level has a boost improvement over nearly all metrics. Among them, the BLEU-4 and CIDEr-D scores of Transformer+quality(2) outperform the baseline 0.7 / 2.6 respectively. With regard to reinforcement training, they achieve BLEU-4 / CIDEr-D scores of 39.5 / 131.4, which surpass the results of baseline models a lot. The experiment shows that the introduction of the quality signal can improve the accuracy.

\begin{table}[tbp]
	\begin{center}
		\begin{tabular}{c|cccccc}
			\hline
			Method & B-1 & B-4 & M & R & C & S \\ \hline
			RL \cite{rennie2017self} & \textbf{80.8} & 39.0 & 29.0 & 58.8 & 129.4 & 22.8 \\ 
			SAT \cite{zhu2021self} & 80.4 & 39.4 & 28.8 & 58.8 & 129.0 & 22.5 \\ 
			+ step 2-5 & 80.6 & \textbf{39.5} & 29.0 & 58.9 & 129.7 & 22.6 \\
			+ step 9 & 80.6 & \textbf{39.5} & 29.1 & \textbf{59.0} & 130.8 & 22.9 \\ 
			Q-SAT & \textbf{80.8} & \textbf{39.5} & \textbf{29.2} & \textbf{59.0} & \textbf{131.4} & \textbf{23.0} \\\hline
		\end{tabular}
	\end{center}
	\caption{Settings and results of ablation studies on the Karpathy test split. Top one result on each metric is in bold.}
\label{Table4}\end{table}

\subsection{Ablation Study}
\label{sec4.4}
In the ablation study, we select Transformer \cite{vaswani2017attention} as our baseline model and fix the quality level to the highest level. Since our Q-SAT method mainly makes two improvements (step 2-5 and step 9 in Algorithm \ref{alg1}) compared to the original SAT method, we explore their respective roles in this section.

From Table \ref{Table4}, we have the following observations: 1) Since the SAT method is designed for CIC tasks, it is not suitable for the control signal of sentence quality. The model trained with SAT method performs even worse than that trained with conventional reinforcement learning methods. It means that with SAT method, the introduction of control signal of sentence quality has a negative effect, which is opposite to the results of cross entropy training. Therefore, it is necessary to make some targeted improvements for the control signal of sentence quality. 2) By adding the step 2-5, the CIC model improves the CIDEr-D score from 129.0 to 129.7. It proves our motivation that determining the center level for a given image help to reduce subsequent control signal changes in step 9, which improves the overall performance of the CIC model. 3) The SAT method discards the optimization of samples resulting in lower rewards than baseline, which is a significant loss for updating. We add the step 9 to retain the optimization of these sentences, boosting the CIDEr-D score from 129.0 to 130.8 on the SAT baseline. 4) When the two parts are combined together, which is a complete version of Q-SAT method, the CIC model achieves the best performance (131.4 on CIDEr-D and 23.0 on SPICE). The experiments above validate the effectiveness of the targeted improvements in our Q-SAT method.

\begin{table}[tbp]
	\begin{center}
	\resizebox{1\columnwidth}{!}{
		\begin{tabular}{c|cccccc}
			\hline
			Model & B-1 & B-4 & M & R & C & S \\ \hline
			ETA \cite{li2019entangled} & {81.5} & 39.3 & 28.8 & 58.9 & 126.6 & 22.7 \\ 
			AoANet \cite{huang2019attention} & 80.2 & 38.9 & {29.2} & 58.8 & 129.8 & 22.4 \\ 
			M2-T \cite{cornia2020meshed} & 80.8 & 39.1 & {29.2} & 58.6 & 131.2 & 22.6  \\ 
			CAAG \cite{song2021image} & - & 39.4 & {29.5} & \textbf{59.2} & 132.2 & 22.8 \\ 
			RSTNet \cite{zhang2021rstnet} & 81.1 & 39.3 & 29.4 & 58.8 & 133.3 & 23.0  \\ 
			DLCT \cite{luo2021dual} & 81.4 & 39.8 & {29.5} & {59.1} & 133.8 & 23.0  \\ 
            $\mathcal{A}^2$ Transformer \cite{fei2022attention} & 81.5 & 39.8 & \textbf{29.6} & {59.1} & 133.9 & 23.0  \\ 
			\hline
			\hline
			DLCT+Q-SAT & \textbf{81.7} & \textbf{40.0} & {29.4} & \textbf{59.2} & \textbf{134.7} & \textbf{23.2} \\
			\hline
		\end{tabular}}
	\end{center}
	\caption{Comparison with the state of the art on MSCOCO Karpathy split. Top one result on each metric is in bold.}
\label{Table5}\end{table}

\subsection{Comparison with Previous Works}
\label{sec4.5}
Table \ref{Table5} reports the results of our Q-SAT method in comparison with the aforementioned state-of-the-art models, which use captions predicted from a single model and optimization on the CIDEr-D score. As shown in Table \ref{Table5}, DLCT+Q-SAT achieves the best performance in all evaluation metrics except METEOR.  In particular, it advances the current state of the art on CIDEr-D by 0.8 points. Compared with previous works, our model benefits from the control signal of sentence quality, thus generating high-quality descriptions.

\section{Conclusion}
In this paper, we propose a new control signal of sentence quality. It aims to make models aware of the quality level of target sentences and treat them differently. Without additional information from ground truth captions, models  controlled  by  the  highest  quality  level  surpass  baseline models on accuracy-based evaluation metrics. Moreover, we propose a novel reinforcement training method called Quality-oriented Self-Annotated Training. The Q-SAT method is specially designed for the control signal of sentence quality and further enhances the accuracy of CIC models. Extensive experiments validate the effectiveness of our proposed methods.

\bibliographystyle{IEEEbib}
\bibliography{paper}

\begin{thebibliography}{10}

\bibitem{Vinyals2015Show}
Oriol Vinyals, Alexander Toshev, Samy Bengio, and Dumitru Erhan,
\newblock ``Show and tell: A neural image caption generator,''
\newblock in {\em Proceedings of the IEEE conference on computer vision and
  pattern recognition}, 2015, pp. 3156--3164.

\bibitem{xu2015show}
Kelvin Xu, Jimmy Ba, Ryan Kiros, Kyunghyun Cho, Aaron Courville, Ruslan
  Salakhudinov, Rich Zemel, and Yoshua Bengio,
\newblock ``Show, attend and tell: Neural image caption generation with visual
  attention,''
\newblock in {\em International conference on machine learning}. PMLR, 2015,
  pp. 2048--2057.

\bibitem{Lin2014Microsoft}
Tsung-Yi Lin, Michael Maire, Serge Belongie, James Hays, Pietro Perona, Deva
  Ramanan, Piotr Doll{\'a}r, and C~Lawrence Zitnick,
\newblock ``Microsoft coco: Common objects in context,''
\newblock in {\em European conference on computer vision}. Springer, 2014, pp.
  740--755.

\bibitem{cornia2019show}
Marcella Cornia, Lorenzo Baraldi, and Rita Cucchiara,
\newblock ``Show, control and tell: A framework for generating controllable and
  grounded captions,''
\newblock in {\em Proceedings of the IEEE/CVF Conference on Computer Vision and
  Pattern Recognition}, 2019, pp. 8307--8316.

\bibitem{deng2020length}
Chaorui Deng, Ning Ding, Mingkui Tan, and Qi~Wu,
\newblock ``Length-controllable image captioning,''
\newblock in {\em European Conference on Computer Vision}. Springer, 2020, pp.
  712--729.

\bibitem{zhu2021self}
Zhangzi Zhu, Tianlei Wang, and Hong Qu,
\newblock ``Self-annotated training for controllable image captioning,''
\newblock {\em arXiv preprint arXiv:2110.08446}, 2021.

\bibitem{Vedantam2015CIDEr}
Ramakrishna Vedantam, C~Lawrence~Zitnick, and Devi Parikh,
\newblock ``Cider: Consensus-based image description evaluation,''
\newblock in {\em Proceedings of the IEEE conference on computer vision and
  pattern recognition}, 2015, pp. 4566--4575.

\bibitem{rennie2017self}
Steven~J Rennie, Etienne Marcheret, Youssef Mroueh, Jerret Ross, and Vaibhava
  Goel,
\newblock ``Self-critical sequence training for image captioning,''
\newblock in {\em Proceedings of the IEEE conference on computer vision and
  pattern recognition}, 2017, pp. 7008--7024.

\bibitem{vaswani2017attention}
Ashish Vaswani, Noam Shazeer, Niki Parmar, Jakob Uszkoreit, Llion Jones,
  Aidan~N Gomez, {\L}ukasz Kaiser, and Illia Polosukhin,
\newblock ``Attention is all you need,''
\newblock {\em Advances in neural information processing systems}, vol. 30,
  2017.

\bibitem{karpathy2015deep}
Andrej Karpathy and Li~Fei-Fei,
\newblock ``Deep visual-semantic alignments for generating image
  descriptions,''
\newblock in {\em Proceedings of the IEEE conference on computer vision and
  pattern recognition}, 2015, pp. 3128--3137.

\bibitem{papineni2002bleu}
Kishore Papineni, Salim Roukos, Todd Ward, and Wei-Jing Zhu,
\newblock ``Bleu: a method for automatic evaluation of machine translation,''
\newblock in {\em Proceedings of the 40th annual meeting of the Association for
  Computational Linguistics}, 2002, pp. 311--318.

\bibitem{lin2004rouge}
Chin-Yew Lin,
\newblock ``Rouge: A package for automatic evaluation of summaries,''
\newblock in {\em Text summarization branches out}, 2004, pp. 74--81.

\bibitem{denkowski2014meteor}
Michael Denkowski and Alon Lavie,
\newblock ``Meteor universal: Language specific translation evaluation for any
  target language,''
\newblock in {\em Proceedings of the ninth workshop on statistical machine
  translation}, 2014, pp. 376--380.

\bibitem{anderson2016spice}
Peter Anderson, Basura Fernando, Mark Johnson, and Stephen Gould,
\newblock ``Spice: Semantic propositional image caption evaluation,''
\newblock in {\em European conference on computer vision}. Springer, 2016, pp.
  382--398.

\bibitem{li2019entangled}
Guang Li, Linchao Zhu, Ping Liu, and Yi~Yang,
\newblock ``Entangled transformer for image captioning,''
\newblock in {\em Proceedings of the IEEE/CVF international conference on
  computer vision}, 2019, pp. 8928--8937.

\bibitem{huang2019attention}
Lun Huang, Wenmin Wang, Jie Chen, and Xiao-Yong Wei,
\newblock ``Attention on attention for image captioning,''
\newblock in {\em Proceedings of the IEEE/CVF international conference on
  computer vision}, 2019, pp. 4634--4643.

\bibitem{cornia2020meshed}
Marcella Cornia, Matteo Stefanini, Lorenzo Baraldi, and Rita Cucchiara,
\newblock ``Meshed-memory transformer for image captioning,''
\newblock in {\em Proceedings of the IEEE/CVF conference on computer vision and
  pattern recognition}, 2020, pp. 10578--10587.

\bibitem{song2021image}
Zeliang Song, Xiaofei Zhou, Zhendong Mao, and Jianlong Tan,
\newblock ``Image captioning with context-aware auxiliary guidance,''
\newblock in {\em Proceedings of the AAAI Conference on Artificial
  Intelligence}, 2021, vol.~35, pp. 2584--2592.

\bibitem{zhang2021rstnet}
Xuying Zhang, Xiaoshuai Sun, Yunpeng Luo, Jiayi Ji, Yiyi Zhou, Yongjian Wu,
  Feiyue Huang, and Rongrong Ji,
\newblock ``Rstnet: Captioning with adaptive attention on visual and non-visual
  words,''
\newblock in {\em Proceedings of the IEEE/CVF conference on computer vision and
  pattern recognition}, 2021, pp. 15465--15474.

\bibitem{luo2021dual}
Yunpeng Luo, Jiayi Ji, Xiaoshuai Sun, Liujuan Cao, Yongjian Wu, Feiyue Huang,
  Chia-Wen Lin, and Rongrong Ji,
\newblock ``Dual-level collaborative transformer for image captioning,''
\newblock in {\em Proceedings of the AAAI Conference on Artificial
  Intelligence}, 2021, vol.~35, pp. 2286--2293.

\bibitem{fei2022attention}
Zhengcong Fei,
\newblock ``Attention-aligned transformer for image captioning,''
\newblock in {\em Proceedings of the AAAI Conference on Artificial
  Intelligence}, 2022, vol.~36, pp. 607--615.

\end{thebibliography}

\end{document}